%% file: main.tex
\documentclass[10pt,twocolumn,letterpaper]{article}

\usepackage[pagenumbers]{cvpr} 


\usepackage{wrapfig}
\usepackage[utf8]{inputenc} 
\usepackage[T1]{fontenc}    
\usepackage{url}            
\usepackage{booktabs}       
\usepackage{amsfonts}       
\usepackage{nicefrac}       
\usepackage{microtype}      
\usepackage{multirow}
\usepackage[table,xcdraw]{xcolor}
\usepackage{makecell}
\usepackage{caption}

\usepackage{amsmath}

\usepackage{graphicx}

\definecolor{cvprblue}{rgb}{0.21,0.49,0.74}
\usepackage[pagebackref,breaklinks,colorlinks,allcolors=cvprblue]{hyperref}

\def\confName{CVPR}
\def\confYear{2026}

\title{CogAD: \underline{Cog}nitive-Hierarchy Guided End-to-End \underline{A}utonomous \underline{D}riving}

\author{
  Zhennan Wang$^{\star}$  \qquad  Jianing Teng$^{\star}$  \qquad  Canqun Xiang  \qquad  Kangliang Chen \\
  Xing Pan \qquad  Lu Deng \qquad  Weihao Gu$^{\dagger}$ \\
  HAOMO.AI Technology Co., Ltd \\
  \tt\small wangzhennan2017@email.szu.edu.cn$^{\star}$, tengjn@alumni.sysu.edu.cn$^{\star}$ \\ 
  \tt\small guwh22@mails.tsinghua.edu.cn$^{\dagger}$
}

\begin{document}
\maketitle
\input{sec/0_abstract}

\footnotetext{$^\star$ \ Equal contribution; $^\dagger$ \ Corresponding author.}

\input{sec/1_intro}

\input{sec/2_related_work}

\input{sec/3_method}

\input{sec/4_experiments}

\input{sec/5_conclusion}
\input{sec/X_suppl}

\newpage
\clearpage
\newpage
\clearpage
\newpage

{
    \small
    \bibliographystyle{ieeenat_fullname}
    \bibliography{main}
}


\end{document}

%% file: sec/0_abstract.tex
\begin{abstract}
    While end-to-end autonomous driving has advanced significantly, prevailing methods remain fundamentally misaligned with human cognitive principles in both perception and planning. 
    In this paper, we propose CogAD, a novel end-to-end autonomous driving model that emulates the hierarchical cognition mechanisms of human drivers. 
    CogAD implements dual hierarchical mechanisms: global-to-local context processing for human-like perception and intent-conditioned multi-mode trajectory generation for cognitively-inspired planning.
    The proposed method demonstrates three principal advantages: comprehensive environmental understanding through hierarchical perception, robust planning exploration enabled by multi-level planning, and diverse yet reasonable multi-modal trajectory generation facilitated by dual-level uncertainty modeling. 
    Extensive experiments on nuScenes and Bench2Drive demonstrate that CogAD achieves state-of-the-art performance in end-to-end planning, exhibiting particular superiority in long-tail scenarios and robust generalization to complex real-world driving conditions.   
\end{abstract}

%% file: sec/1_intro.tex
\section{Introduction}
\label{sec:intro}


In recent years, end-to-end autonomous driving (E2E-AD) has been a topic of interest~\cite{jiang2023vad, hu2023planning, weng2024drive, sun2024sparsedrive, jia2025drivetransformer}. 
E2E-AD methods elevate the performance upper bound of autonomous driving, while broadening the architectural design space for models~\cite{weng2024drive}.
Although significant progress has been made in the field, current approaches exhibit substantial discrepancies from human driving processes in both perception and planning patterns, from the perspective of cognitive psychology.

\begin{figure}[t]
  \centering
  \includegraphics[width=1.0\columnwidth]{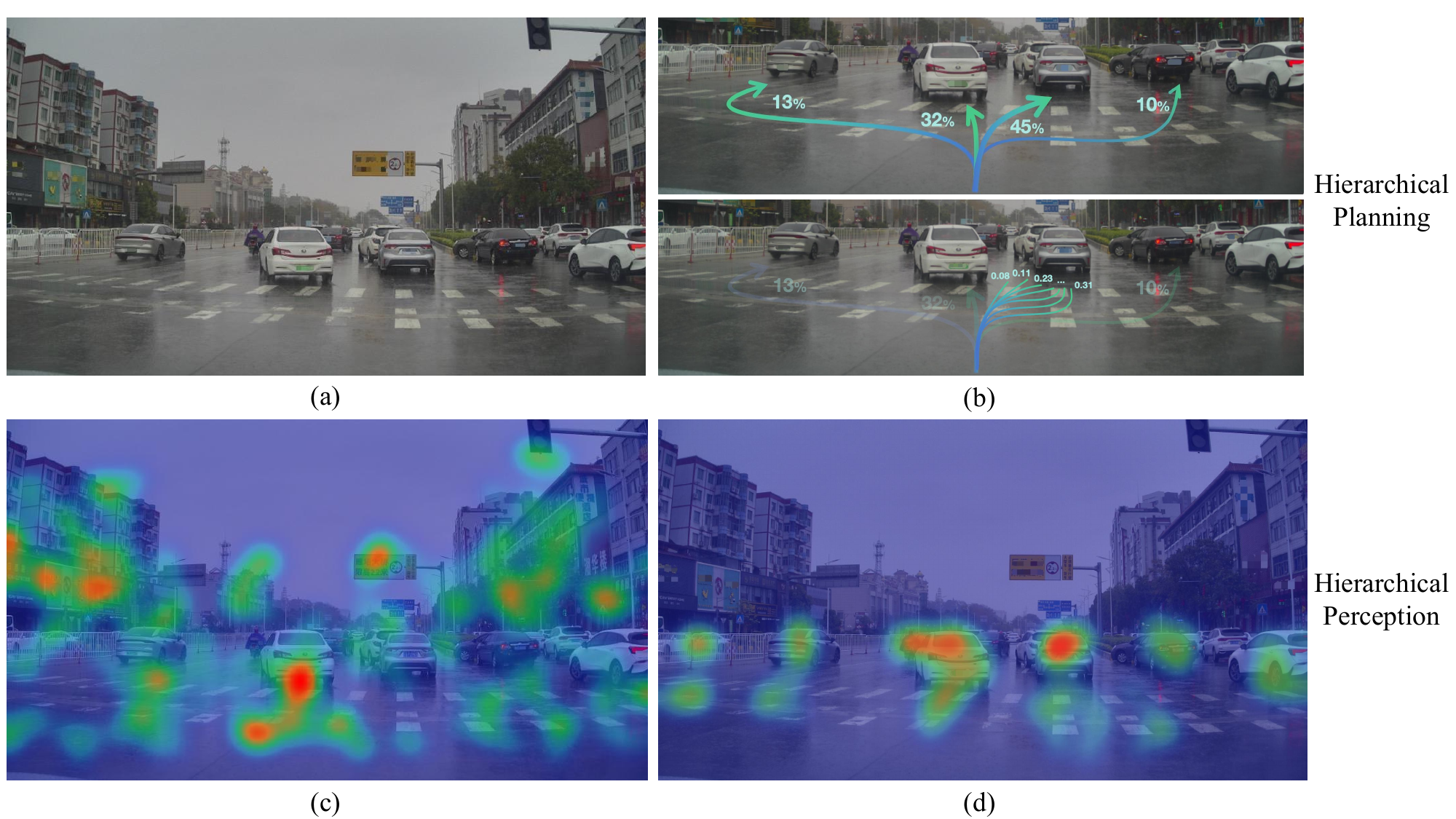}
  \caption{
  Human drivers scan surrounds (c) before focusing on key objects (d), and plan hierarchically from intent to trajectory (b).
  }
  \label{fig:fig1-main}
  \vspace*{-5mm}
\end{figure}

In the perception pattern, we argue that the perception of human drivers is hierarchical. 
Cognitive psychology supports the pre-attentive two-stage theory of human visual perception~\cite{koch1987shifts}, which posits that individuals first perform global scanning of the scene to identify contextual or spatial cues, followed by selective attention allocation to critical local objects for detailed analysis. 
This mechanism aligns with neurobiological evidence~\cite{felleman1991distributed} demonstrating that early visual areas prioritize scene-wide features, while higher-order regions refine task-specific processing through context-dependent modulation. 
In terms of driving environmental perception, human drivers typically initiate their environmental perception by establishing holistic scene comprehension of the traffic scenario, encompassing all contextual elements as illustrated in \cref{fig:fig1-main} (c),  
followed by selective attention prioritization towards critical dynamic objects such as proximal vehicles, pedestrians, and lane markings, as illustrated in \cref{fig:fig1-main} (d).

In the planning pattern, we argue that the planning of human drivers is hierarchical. The BDI (Belief-Desire-Intention) cognitive  model~\cite{bratman1987intention, rao1995bdi} decomposes human action planning into hierarchically organized processes, distinguishing between high-level intentional states (e.g., goal setting) and low-level behavioral executions (e.g., motor sequence implementation).
In terms of driving planning, human drivers typically follow a hierarchical planning process: intentional planning at the higher level (e.g., lane change) followed by trajectory planning at the lower level (e.g., generating a specific trajectory). 
For instance, drivers first establish a global intent, which then constrains the generation of local trajectories, as illustrated in \cref{fig:fig1-main} (b).

From the perspective of hierarchical cognitive psychology, recent methods like~\cite{weng2024drive, jiang2023vad, hu2023planning} fail to 
incorporate hierarchical perception, while others like~\cite{sun2024sparsedrive, jia2025drivetransformer, liao2024diffusiondrive} lack hierarchical planning, as detailed in \cref{sec:rltwork}. 
These analyses highlight a critical gap in aligning E2E-AD models with human drivers' cognitive principles. 
To bridge this gap, we propose CogAD, an end-to-end autonomous driving model that emulates human hierarchical cognition. 
For hierarchical perception, we design a sequential interaction mechanism where the ego-vehicle first processes global BEV features to capture environmental context, then focuses on critical instance-level elements. 
For hierarchical planning, CogAD first plans high-level driving intents, then generates corresponding low-level trajectories, mirroring human multi-level planning.

The advantages of CogAD are summarized as follows: 
(a) CogAD captures hierarchical environmental representations, encompassing both holistic scene features and critical element attributes; 
(b) CogAD achieves comprehensive planning space exploration through dual-level uncertainty modeling at both intent and trajectory levels; 
(c) CogAD inherently supports multi-modal trajectory planning, generating diverse and plausible trajectories for downstream tasks. 
In summary, the main contributions of this paper are three-fold:
\begin{itemize}
    \item We propose a hierarchical scene-instance perception paradigm that significantly enhances the ego vehicle's scene understanding capabilities.
    \item We develop a hierarchical intent-trajectory planning mechanism that simultaneously enhances both behavioral diversity and motion rationality in end-to-end autonomous driving. 
    \item CogAD achieves state-of-the-art performance in both open-loop and closed-loop evaluation, with particularly significant improvements in long-tail scenarios compared to prior methods.
\end{itemize}

%% file: sec/2_related_work.tex
\section{Related Work}
\label{sec:rltwork}

Significant progress has been made in end-to-end autonomous driving (E2E-AD) approaches, pioneered by prior works such as~\cite{bojarski2016end, codevilla2018end, pomerleau1988alvinn}. 

Existing approaches exhibit distinct limitations in hierarchical feature utilization. 
Scene-centric methods~\cite{chitta2022transfuser, li2024ego, li2024hydra, weng2024drive, xing2025goalflow} emphasize scene-level features while systematically overlooking instance-level characteristics.  
Specifically, ParaDrive~\cite{weng2024drive} and BEV-Planner~\cite{li2024ego} solely employ BEV features extracted from visual inputs, whereas TransFuser-based architectures~\cite{chitta2022transfuser, li2024hydra, xing2025goalflow} integrate BEV features obtained through multimodal lidar-camera fusion. 
Conversely, instance-oriented approaches~\cite{su2024difsd, doll2024dualad, jiang2023vad, sun2024sparsedrive, li2024enhancing, zheng2024genad} prioritize instance-specific attributes but fail to integrate scene-level contextual information. 
For instance, VAD~\cite{jiang2023vad} establishes interactions between ego queries and map/agent queries without BEV feature integration, while SparseDrive~\cite{sun2024sparsedrive} restricts feature interactions to current and historical instance-level representations. 
Although DiFSD~\cite{su2024difsd} implements hierarchical interactions within its planning module, these remain confined to inter-instance communication. 
Despite leveraging both scene-level context and instance-level details, methods like~\cite{hu2023planning, chen2024ppad, jia2025drivetransformer, ye2023fusionad} do not exhibit a clear perceptual hierarchy in feature interaction. 
Notably, UniAD~\cite{hu2023planning}, FusionAD~\cite{ye2023fusionad}, and PPAD~\cite{chen2024ppad} adopt an inverted interaction sequence—prioritizing instance-level processing before BEV feature integration—contrary to human cognitive patterns. 
DriveTransformer~\cite{jia2025drivetransformer} employs iterative refinement between instance and scene representations, yet lacks explicit hierarchical separation. 
In contrast, CogAD establishes a cognitively-inspired perception hierarchy through a scene-prioritized interaction paradigm followed by instance-level refinement, achieving principled decoupling of hierarchical representations. 

\begin{figure*}[t]
  \centering
  \includegraphics[width=0.8\linewidth]{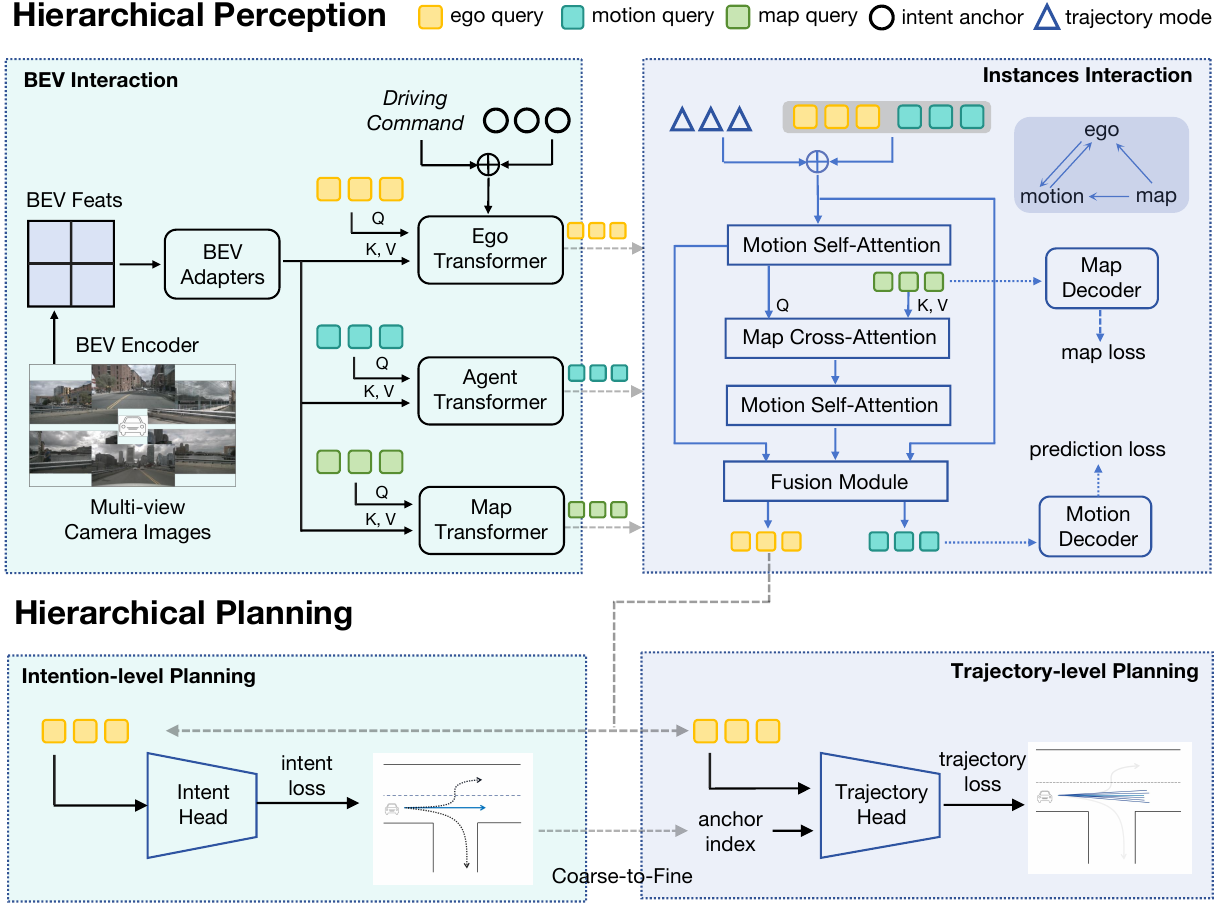}
  \caption{The overall framework of CogAD. 
  CogAD extracts BEV features into task-specific queries, then performs cross-task instance feature interaction, forming a hierarchical perception paradigm. 
  Meanwhile, CogAD implements intent-level planning and subsequently conducts trajectory-level planning, establishing a hierarchical planning mechanism.
  }
  \label{fig:framework}
  \vspace*{-5mm}
\end{figure*}

Recent research in E2E-AD reveals notable limitations in hierarchical planning with uncertainty modeling across different abstraction levels. 
Current approaches predominantly focus on deterministic trajectory generation~\cite{hu2022st, hu2023planning, jiang2023vad, weng2024drive, li2024ego, sadat2020perceive, chen2024ppad, chitta2022transfuser, li2024enhancing, zheng2024genad, ye2023fusionad, guo2024end, doll2024dualad, shen2025divide}, fundamentally neglecting uncertainty quantification. 
While multimodal trajectory prediction methods~\cite{sun2024sparsedrive, tang2025hip, jia2025drivetransformer, su2024difsd, zhang2024sparsead} incorporate trajectory-level uncertainty, they fail to address higher-order intent uncertainty. 
Notably, SparseDrive~\cite{sun2024sparsedrive} employs learnable embeddings to represent the planning modes and adopts the winner-takes-all strategy for loss computation, analogous to our trajectory uncertainty modeling. 
HiP-AD~\cite{tang2025hip} implements temporal-spatial hierarchical planning with driving-style modeling but lacks explicit intent representation. 
Alternative approaches~\cite{wang2024driving, liao2024diffusiondrive} extract diverse motion patterns that capture intent-level uncertainty, neglecting trajectory-level uncertainty. 
In particular, Drive-WM~\cite{wang2024driving} employs human-specified high-level commands to represent intent in a heuristic manner. 
Similar to our intent uncertainty module, DiffusionDrive~\cite{liao2024diffusiondrive} constructs planning anchors by offline K-Means clustering on the training set, yet omit trajectory-level uncertainty. 
Other methods generating multimodal trajectories via large-scale sampling~\cite{chen2024vadv2, li2024hydra, xing2025goalflow} conflate intent and trajectory uncertainties. 
Some non-end-to-end autonomous driving approaches also incorporate hierarchical planning design. 
Methods like~\cite{kim2022hierarchical, qi2022hierarchical} adopt a temporal hierarchical paradigm, diverging fundamentally from our uncertainty-aware intent-trajectory dual hierarchy. 
Other approaches proposed in~\cite{wang2020learning, van2020hierarchical, du2023hierarchical, lu2020hierarchical} similarly employ hierarchical planning, yet critically depend on manually engineered command definitions rather than data-driven representations. 
Contemporary LLM-based planners~\cite{tian2024drivevlm, huang2024drivlme, mei2024continuously, chen2024asynchronous, tian2024tokenize, jiang2024senna, qian2024fasionad, xu2024vlm, guo2025vdt} demonstrate conceptual parallels through meta-action prediction but suffer from manually defined commands and computational inefficiency. 
Distinct from these approaches, our CogAD introduces a novel bi-level uncertainty-aware planning mechanism that decouples intent uncertainty from trajectory uncertainty, achieving superior performance with optimized computational efficiency.

%% file: sec/3_method.tex
\section{Method}



\subsection{Overview} 


As depicted in \cref{fig:framework}, the proposed CogAD implements a unified multi-task architecture that integrates perception, prediction, and planning in a coordinated manner. 
To comprehensively capture environmental information, we design a hierarchical feature interaction framework. 
First, we establish interactions between task-specific instance queries (ego, agent, and map) and dense Bird's Eye View (BEV) features while incorporating intent uncertainty in the ego-vehicle's planning queries. 
Subsequently, cross-task instance interaction is conducted among task-specific sparse representations with inter-dependencies, where trajectory uncertainty is explicitly modeled in the ego-vehicle's planning instance.
Next, we introduce the hierarchical BEV-instance perception paradigm and the hierarchical intent-trajectory planning mechanism.

\subsection{Hierarchical Perception}

\paragraph{Interaction with BEV Features.} 

BEV features provide a unified scene representation that captures global context and geometric relationships. 
We therefore initiate the feature interaction process by bridging the task-specific instance queries with BEV features. 
For BEV features extraction, we implement the BEVFormer~\cite{li2022bevformer} framework due to its effective spatial-temporal fusion mechanism, which captures both geometric layouts and temporal motion dynamics. 
We develop a dedicated BEV adapter for each task to enhance the task-specific adaptability of BEV features. 
\begin{equation}\label{equ:plan_bev}
    \begin{split}
        Q_{ego}&=E_{ego} + E_{cmd} + E_{intent} \\
        I_{ego}&=Transformer(Q=Q_{ego}, \\ & \hspace{2.3cm} K,V=Adapter_{ego}(F_{bev}))
    \end{split}
\end{equation}
\cref{equ:plan_bev} models the interaction between planning queries and BEV features $F_{bev}$. 
$Transformer$ represents the standard transformer module~\cite{vaswani2017attention}. 
CogAD integrates ego embedding $E_{ego}$, high-level driving commands (e.g., lane-keeping/turning) $E_{cmd}$, and intent anchors $E_{intent}$ to initialize the ego-vehicle queries $Q_{ego}$, which are then fed into the Ego Transformer module shown in \cref{fig:framework}. 
The intent anchors encode the ego vehicle's coarse planning space, with implementation details provided in \cref{sec:planning}. 

CogAD also performs online mapping and dynamic obstacle detection via the interaction between BEV features and map/motion queries, constituting mission-critical modules that are algorithmically coupled with the ego planning task. 
Online mapping provides topological constraints for planning.  
CogAD employs multiple learnable spatial interaction queries to decode structured road topology from BEV features through multi-head cross attention, implemented through the MapTransformer module in \cref{fig:framework}. 
This interaction process can be formally defined as: 
\begin{equation}\label{equ:map_bev}
\begin{split}
  I_{map}&=Transformer(Q=Q_{map}, \\
  &\hspace{2.3cm}  K,V=Adapter_{map}(F_{bev}))
\end{split}
\end{equation}
Obstacle detection provides precise spatial localization of surrounding agents, thereby ensuring safety-critical planning. 
CogAD initiates with interaction queries that extract spatiotemporal information from BEV features via the DetTransformer module in \cref{fig:framework}, whereby geometric attributes (position, heading angle) and semantic properties (classification confidence) are jointly predicted via a lightweight multi-layer perceptron. 
The interaction process can be formulated as follows:
\begin{equation}\label{equ:agent_bev}
\begin{split}
  I_{agent}&=Transformer(Q=Q_{agent}, \\
  &\hspace{2.3cm}  K,V=Adapter_{agent}(F_{bev}))
\end{split}
\end{equation}

\begin{table*}[!t]
\centering
\caption{Planning performance on nuScenes dataset. We use * for LLM-based method.}
\label{tab:performance}
\begin{tabular}{lllllllllc}
\midrule
                                               & \multicolumn{4}{c}{L2 (m) $\downarrow$}                                                                                                                                          & \multicolumn{4}{c}{Collision (\%) $\downarrow$}                                                                                                                                  & \multicolumn{1}{l}{}                              \\ \cmidrule(lr){2-5} \cmidrule(lr){6-9}
\multirow{-2}{*}{Method}                       & 1s                          & 2s                          & 3s                          & Avg.                                                                     & 1s                          & 2s                          & 3s                          & Avg.                                                                     & \multirow{-2}{*}{\begin{tabular}[c]{@{}c@{}}Latency\\ (ms)\end{tabular}} \\ \midrule
UniAD ~\cite{hu2023planning}                                         & 0.48                        & 0.74                        & 1.07                        & \multicolumn{1}{l|}{0.76}                                                & 0.12                        & 0.13                        & 0.28                        & \multicolumn{1}{l|}{0.17}                                                & 555.6                                             \\
VAD ~\cite{jiang2023vad}                                       & 0.41                        & 0.70                        & 1.05                        & \multicolumn{1}{l|}{0.72}                                                & 0.07                        & 0.17                        & 0.41                        & \multicolumn{1}{l|}{0.22}                                                & 224.3                                             \\
ParaDrive ~\cite{weng2024drive}                                            & 0.25                        & 0.46                        & 0.74                        & \multicolumn{1}{l|}{0.48}                                                & 0.14                        & 0.23                        & 0.39                        & \multicolumn{1}{l|}{0.25}                                                & -                                             \\
SparseDrive ~\cite{sun2024sparsedrive}                                            & 0.29                        & 0.55                        & 0.91                        & \multicolumn{1}{l|}{0.58}                                                & 0.01                        & 0.02                        & 0.13                        & \multicolumn{1}{l|}{0.06}                                                & 136.9                                             \\
DiffusionDrive ~\cite{liao2024diffusiondrive}                                  & 0.27                        & 0.54                        & 0.90                         & \multicolumn{1}{l|}{0.57}                                                & 0.03                        & 0.05                        & 0.16                        & \multicolumn{1}{l|}{0.08}                                                & -                                                 \\
\midrule
Senna* ~\cite{jiang2024senna}                                         & 0.37                        & 0.54                        & 0.86                        & \multicolumn{1}{l|}{0.59}                                                & 0.09                        & 0.12                        & 0.33                        & \multicolumn{1}{l|}{0.18}                                                & -                                                 \\
TOKEN* ~\cite{tian2024tokenize}                                         & 0.26                        & 0.71                        & 1.47                        & \multicolumn{1}{l|}{0.81}                                                & -                           & -                           & -                           & \multicolumn{1}{l|}{-}                                                   & -                                                 \\
VLM-AD* ~\cite{xu2024vlm}                                         & 0.30                        & 0.54                        & 0.80                        & \multicolumn{1}{l|}{0.55}                                                & 0.11                        & 0.15                        & 0.38                        & \multicolumn{1}{l|}{0.21}                                                & -                                                 \\ \midrule
CogAD                                          & 0.24                        & 0.45                        & 0.74                        & \multicolumn{1}{l|}{0.48}                                                & 0.00                        & 0.02                        & 0.15                        & \multicolumn{1}{l|}{0.06}                                                & 92.9                                              \\ \midrule

\end{tabular}
\vspace{-5mm}
\end{table*}

\vspace{-5mm}
\paragraph{Cross-Task Instance Interaction.}
Although BEV features facilitate a unified scene understanding, they exhibit limitations in representing the multimodal future trajectories of dynamic agents and ego-vehicle. 
Additionally, gradient conflicts~\cite{chen2025m3net, huang2023fuller} are likely to arise among the loss functions of multiple tasks, within the BEV space. 
To address these limitations, CogAD involves inter-instances interactions across different tasks, as shown in the upper right part of \cref{fig:framework}.  

In the planning module, we account for trajectory uncertainty by introducing motion mode embeddings, as elaborated in \cref{sec:planning}.
Regarding obstacles, we enhance their representations by incorporating dynamic attributes through motion prediction, aiming to avoid potential collisions with the ego-vehicle's planned trajectory. 
Similarly, we introduce motion mode embeddings to facilitate multi-modal motion prediction. 
We align with the philosophy advocated by SparseDrive~\cite{sun2024sparsedrive}, which emphasizes that motion prediction and planning should account for bidirectional interactions. 
To this end, the instance queries for ego-vehicle planning and motion prediction are concatenated and then fed into a self-attention (SA) module, thereby facilitating bidirectional information interaction between the two tasks. 
This design enables the planning task to consider potential future behaviors of surrounding agents, while the motion prediction task also benefits from the planned trajectory of the ego-vehicle. 
This interaction process can be formulated as follows: 
\begin{equation}\label{equ:plan_agent}
    \begin{split}
        I_{ego} = I_{ego} + E_{mode}, \hspace{6mm}
        I_{mot} = I_{agent} + E_{mode} \\
        I_{ego}^{'}, I_{mot}^{'} = Transformer(Q,K,V=[I_{ego}, I_{mot}])
    \end{split}
\end{equation}
Moreover, given the critical role of map information in both tasks (e.g., lane-keeping assistance, identifying road curvature, and providing critical boundary information), CogAD employs cross-attention (CA) to separately distill map features into the instance embeddings of each task. 
After acquiring map information, both the planning and motion prediction instance embeddings update their respective feature representations. 
To ensure mutual awareness of potential future trajectories between the ego-vehicle and surrounding agents, CogAD employs self-attention to model bidirectional interactions based on the updated embeddings. 
\begin{equation}\label{equ:plan_agent}
\begin{split}
I_{ego}^{''}, I_{mot}^{''} = Transformer(Q,K,V= \hspace{1.8cm}  \\
        Transformer(Q=[I_{ego}^{'}, I_{mot}^{'}], K,V=I_{map})) 
\end{split}
\end{equation}
Meanwhile, to guarantee effective utilization of multi-stage features and enhance the absorption of supervised signals, we adopt skip connections that directly transport the outputs from the BEV interaction ($I_{ego}, I_{mot}$), the initial ego-agent self-attention interaction ($I_{ego}^{'}, I_{mot}^{'}$), and the final instance embeddings ($I_{ego}^{''}, I_{mot}^{''}$) to the corresponding task heads. 

\begin{figure}[t]
  \centering
  \includegraphics[width=1.0\columnwidth]{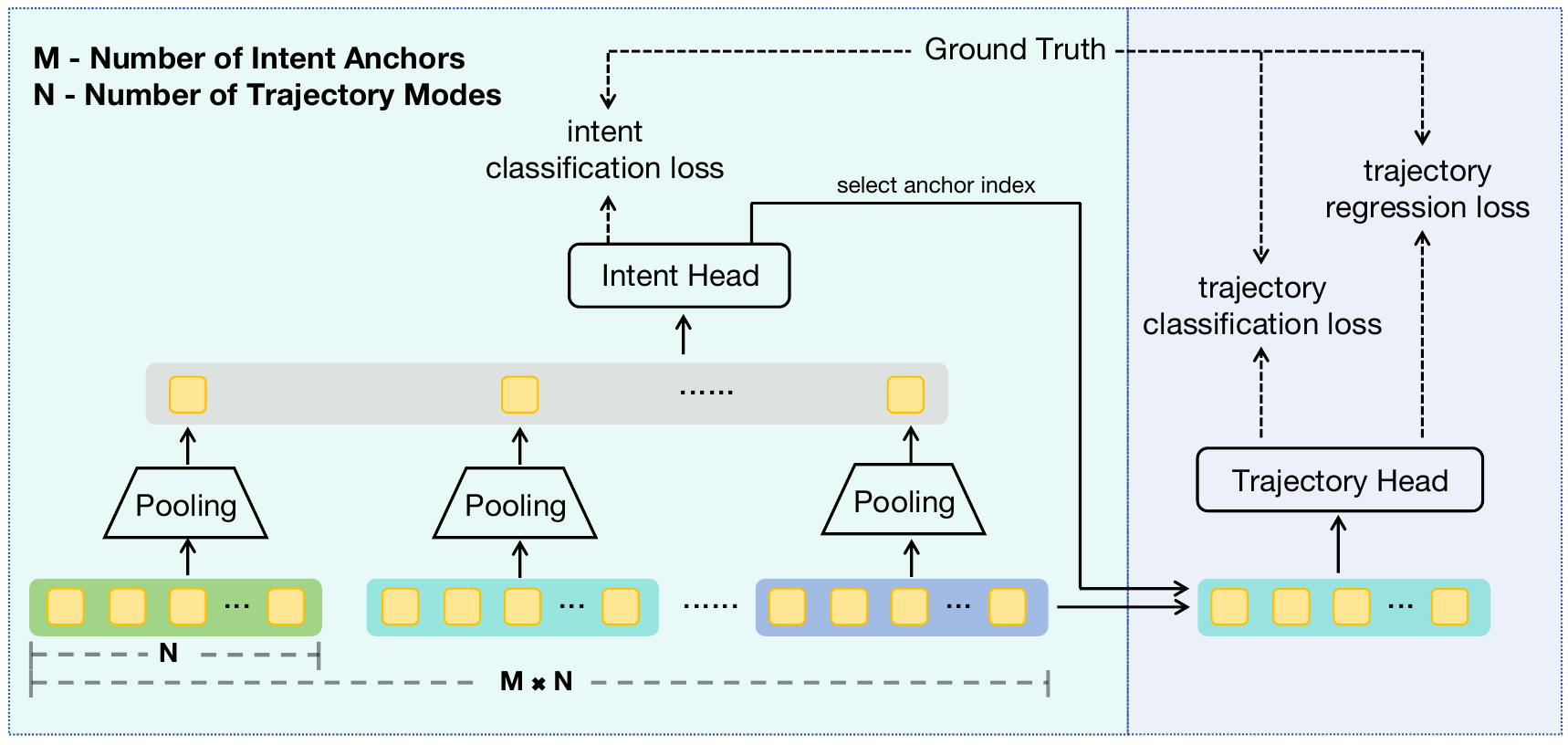}
  \caption{Intent-level and Trajectory-level uncertainty modeling.}
  \label{fig:uncertainty}
  \vspace*{-5mm}
\end{figure}

\subsection{Hierarchical Planning}
\label{sec:planning}

\paragraph{Intent Uncertainty.} 
Intent uncertainty is introduced during the interaction process between the ego-vehicle and the BEV features. 
Intent uncertainty characterizes the inherent ambiguity associated with the high-level intent or targeted objective of an autonomous vehicle in dynamic environments. 
In driving scenarios, this can involve uncertainty regarding the lane-change intentions or trajectory convergence points toward which the ego-vehicle is actively steering. 
Modeling intent uncertainty is crucial for precisely forecasting the ego-vehicle's future behavior and improving planning robustness under uncertain conditions~\cite{chai2019multipath}. 

Unlike many LLM-based autonomous driving methods~\cite{tian2024drivevlm, jiang2024senna} that rely on hand-crafted meta-actions to represent intent, CogAD leverages a data-driven approach to obtain intent points. 
Following practices similar to those in VADv2~\cite{chen2024vadv2} and Hydra-MDP~\cite{li2024hydra}, we employ K-means clustering to generate trajectory anchors, which serve as intent prototypes.  
To simplify the implementation, we utilize online K-means for constructing trajectory anchors.
Notably, as our focus is on high-level intent discovery rather than fine-grained trajectory clustering, the cluster count is significantly lower than these previous methods~\cite{chen2024vadv2, li2024hydra}. 
After the hierarchical perception, the ego queries are pooled along the anchor dimension and subsequently processed through the intent head to compute the intent classification loss or derive the selected anchor index, as illustrated in the left part of \cref{fig:uncertainty}.

\paragraph{Trajectory Uncertainty.} 
Trajectory uncertainty is incorporated into the interaction process between planning instances and motion prediction instances. 
Trajectory uncertainty, which refers to the unpredictability of a vehicle's future path due to factors such as environmental dynamics, diverse behaviors of road users, and subjective driving behavior, exists in both planning and motion prediction tasks. 
From an alternative perspective, trajectory uncertainty can be conceptualized as the uncertainty pertaining to the offset from the trajectory anchors within the latent space. 

Drawing inspiration from MultiPath++~\cite{varadarajan2022multipath++}, our hypothesis posits that trajectory uncertainty follows some intrinsic motion patterns. 
To capture these motion patterns, we directly employ learnable mode embeddings to represent them. 
We argue that these motion patterns demonstrate consistency across both planning and motion prediction tasks, given that both tasks fundamentally involve forecasting the future trajectories of agents. 
Therefore, we utilize shared mode embeddings for both planning and motion prediction tasks. 
Our experimental results indicate that this strategy not only enhances model performance, but also effectively avoids mode collapse~\cite{chai2019multipath, rhinehart2018r2p2, hong2019rules, liao2024diffusiondrive}. 
Finally, based on the anchor selected by the intent head, the trajectory head generates multi-mode trajectories. 
During training, a winner-takes-all strategy~\cite{varadarajan2022multipath++, lee2016stochastic} is employed to compute the regression loss and classification loss, as shown on the right part of \cref{fig:uncertainty}.

\subsection{Training and Inference}
\paragraph{Training.}
The system enables joint optimization of all task objectives via differentiable end-to-end learning. 
In particular, the planning module incorporates two core loss components: an intent grounding loss corresponding to the intent uncertainty and a multi-mode trajectory imitation loss corresponding to the trajectory uncertainty. 
For intent grounding, the intent anchor that exhibits minimum spatial deviation from the ground truth ego trajectory is designated as the label. 
To model the multi-mode imitation behavior, 
we employ the winner-takes-all strategy~\cite{lee2016stochastic} to select the prediction closest to the ground-truth, following motion prediction paradigms~\cite{varadarajan2022multipath++, jiang2023vad}. 
In addition, we incorporate the planning constraints utilized in VAD~\cite{jiang2023vad}. 
The overall training loss is the weighted sum of the aforementioned components: 
\begin{equation}\label{equ:loss}
\begin{split}
    \mathcal{L} = &\lambda_1 \mathcal{L}_{\text{map}} + \lambda_2 \mathcal{L}_{\text{det}} + \lambda_3 \mathcal{L}_{\text{mot}} + \lambda_4 \mathcal{L}_\text{plan\_intent} + \\
    &\lambda_5 \mathcal{L}_\text{plan\_wta} +\lambda_6 \mathcal{L}_\text{plan\_constr} +\lambda_7 \mathcal{L}_\text{k-means}
\end{split}
\end{equation}
where $\mathcal{L}_{\text{map}}$ denotes the online mapping loss, $\mathcal{L}_{\text{det}}$ denotes the obstacle detection loss, $\mathcal{L}_{\text{mot}}$ denotes the motion prediction loss, $\mathcal{L}_\text{plan\_intent}$ denotes the intent grounding loss, $\mathcal{L}_\text{plan\_wta}$ denotes the multi-mode trajectory imitation loss, $\mathcal{L}_\text{plan\_constr}$ denotes the overall planning constraints, and the $\mathcal{L}_\text{k-means}$ drives the online K-means clustering to generate intent anchors. 
\cref{equ:loss} is optimized by joint training all CogAD components, with the loss weighting parameters set to $\lambda_1=\lambda_2=2.0$, $\lambda_3=0.2$, and $\lambda_4=\lambda_5=\lambda_6=\lambda_7=1.0$.

\begin{table}[t]
\centering
\caption{Long-tail scenario performance comparison.}
\label{tab:performance-long-tail}
\setlength{\tabcolsep}{4pt}
\begin{tabular}{lllllc}
\toprule
                         & \multicolumn{4}{c}{L2 (m) $\downarrow$}                               & CR (\%) $\downarrow$ \\ 
                         \cmidrule(lr){2-5} \cmidrule(lr){6-6} 
\multirow{-2}{*}{Method} & 1s            & 2s            & 3s            & Avg.        & Avg.          \\ \midrule
\multicolumn{6}{c}{\cellcolor[HTML]{EFEFEF}\textbf{3-point turn (zero-shot)}}                             \\
VAD~\cite{jiang2023vad}                 & 0.71          & 1.66          & 3.24          & 1.57          & 0.00           \\
PARA-Drive~\cite{weng2024drive}               & 0.50          & 1.38          & 2.76          & 1.29          & 5.33           \\
TOKEN~\cite{tian2024tokenize}                    & 0.39          & 1.29          & 2.60          & 1.18          & 4.00           \\
DiMA~\cite{hegde2025distilling}                     & \textbf{0.36} & 1.18          & 2.37          & 1.05          & 0.00           \\
CogAD                      & 0.41          & \textbf{0.81} & \textbf{1.47} & \textbf{0.90} & \textbf{0.00}  \\
\multicolumn{6}{c}{\cellcolor[HTML]{EFEFEF}\textbf{Resume from stop}}                                     \\
VAD~\cite{jiang2023vad}                 & 0.60          & 1.72          & 2.83          & 1.42          & 0.00           \\
PARA-Drive~\cite{weng2024drive}                & 0.14          & 0.79          & 2.30          & 0.85          & 0.00           \\
TOKEN~\cite{tian2024tokenize}                    & 0.13          & 0.70          & 1.58          & 0.65          & 0.00           \\
DiMA~\cite{hegde2025distilling}          & 0.15          & 0.65          & 1.34          & 0.66          & 0.00           \\
CogAD                      & \textbf{0.10} & \textbf{0.27} & \textbf{0.64} & \textbf{0.34} & \textbf{0.00}  \\
\multicolumn{6}{c}{\cellcolor[HTML]{EFEFEF}\textbf{Overtake}}                                             \\
VAD~\cite{jiang2023vad}                  & 0.46          & 1.16          & 2.17          & 1.06          & 2.49           \\
PARA-Drive~\cite{weng2024drive}               & 0.27          & 0.89          & 1.94          & 0.85          & 2.30           \\
TOKEN~\cite{tian2024tokenize}                      & 0.29          & 0.77          & 1.63          & 0.74          & \textbf{0.00}  \\
DiMA~\cite{hegde2025distilling}                     & \textbf{0.24}          & 0.72          & 1.50          & 0.66          & 1.29           \\
CogAD                      & 0.28          & \textbf{0.53} & \textbf{0.86} & \textbf{0.56} & 0.28           \\ \midrule
\end{tabular}
\vspace{-21pt}
\end{table}

\paragraph{Inference.}


Benefiting from our hierarchical uncertainty modeling, the inference stage supports multiple operational modes. 
(a) Deterministic Trajectory: Select the highest-confidence intent and trajectory at both levels.
(b) Intent Sampling: Apply probabilistic sampling at the intent level while selecting the highest-confidence trajectory at the trajectory level.
(c) Trajectory Sampling: Choose the highest-confidence intent and apply probabilistic sampling at the trajectory level.
(d) Dual-Level Sampling: Employ probabilistic sampling at both intent and trajectory levels.
The probabilistic sampling mechanism enhances trajectory diversity, which is critical in closed-loop experiments to mitigate the risk of the ego vehicle becoming trapped in suboptimal positions.

%% file: sec/4_experiments.tex
\section{Experiments}
\label{exp}


Our experiments are conducted on the nuScenes dataset~\cite{caesar2020nuscenes} and the Bench2Drive benchmark~\cite{jia2024bench2drive} for open-loop and closed-loop evaluations, respectively. 
We follow TOKEN~\cite{tian2024tokenize} and manually construct a long-tail scenario validation set curated from nuScenes for comprehensive evaluation, including three scenarios: 1) executing 3-point turns; 2) resuming motion after a full stop; 3) overtaking parked cars through the oncoming lane. 
CogAD intentionally excludes any form of ego state or historical trajectory as input to avoid overfitting~\cite{li2024ego, zhai2023rethinking}. 
Moreover, we set the number of intent anchors to 30 and the trajectory modes to 6. 
For fairness, we compare only against methods that do not utilize this information and adhere to the same training protocol of 60 epochs on nuScenes and 6 epochs on Bench2Drive.

\begin{table}[t]
\caption{Closed-loop  evaluation results on Bench2Drive. DS: Driving Score. SR: Success Rate. }
\label{tab:b2d1}
\centering
\begin{tabular}{llcc}
\toprule
\multicolumn{1}{l}{Method}              & DS $\uparrow$        & \multicolumn{1}{l}{SR (\%) $\uparrow$}          & \multicolumn{1}{l}{Latency(ms)} \\ 
\midrule
\multicolumn{1}{l}{UniAD~\cite{hu2023planning}}       & 45.81 & \multicolumn{1}{c}{16.36}  & 663.4                                            \\
\multicolumn{1}{l}{VAD~\cite{jiang2023vad}}              & 42.35 & \multicolumn{1}{c}{15.00}  & 278.3                                            \\
\multicolumn{1}{l}{GenAD~\cite{zheng2024genad}}            & 44.81 & \multicolumn{1}{c}{15.91}      & -                                                \\
\multicolumn{1}{l}{MomAD~\cite{song2025don}}            & 44.54 & \multicolumn{1}{c}{16.71}  & -                                                \\
\multicolumn{1}{l}{CogAD}             & \textbf{48.30} & \multicolumn{1}{c}{\textbf{24.00}}  & \textbf{121.3}                                            \\ \midrule
\end{tabular}
\vspace*{-5mm}
\end{table}

\subsection{Main Results}

\begin{table*}[!t]
\centering
\vspace{-5pt}
\caption{Ablation for designs in Hierarchical Perception and Hierarchical Planning modules.}
\vspace{-5pt}
\label{tab:ablation1}
\begin{tabular}{cc@{\hspace{3pt}}c|l@{\hspace{9pt}}l@{\hspace{9pt}}l@{\hspace{9pt}}l|l@{\hspace{9pt}}l@{\hspace{9pt}}l@{\hspace{9pt}}l}
\toprule
\multicolumn{3}{c|}{Ablation Setting}                                                                                                                                                                                                         & \multicolumn{4}{c|}{L2 (m) $\downarrow$ }                       & \multicolumn{4}{c}{Collision (\%) $\downarrow$}                \\ \midrule
\multicolumn{1}{c|}{}                                                                                     & \begin{tabular}[c]{@{}c@{}}BEV \\ Ineraction\end{tabular}     & \begin{tabular}[c]{@{}c@{}}Instance \\ Interaction\end{tabular}   & 1s   & 2s   & 3s   & \cellcolor[HTML]{C0C0C0}Avg. & 1s   & 2s   & 3s   & \cellcolor[HTML]{C0C0C0}Avg. \\
\multicolumn{1}{c|}{}                                                                                     & -                                                             & $\surd$                                                                 & 0.27 & 0.49 & 0.81 & \cellcolor[HTML]{C0C0C0}0.52 & 0.05 & 0.11 & 0.23 & \cellcolor[HTML]{C0C0C0}0.13 \\
\multicolumn{1}{c|}{}                                                                                     & $\surd$                                                             & -                                                                 & 0.26 & 0.45 & 0.75 & \cellcolor[HTML]{C0C0C0}0.49 & 0.03 & 0.09 & 0.23 & \cellcolor[HTML]{C0C0C0}0.12 \\
\multicolumn{1}{c|}{\multirow{-4}{*}{\begin{tabular}[c]{@{}c@{}}Hierarchical \\ Perception\end{tabular}}} & $\surd$                                                             & $\surd$                                                                 & 0.24 & 0.45 & 0.74 & \cellcolor[HTML]{C0C0C0}\textbf{0.48} & 0.00 & 0.02 & 0.15 & \cellcolor[HTML]{C0C0C0}\textbf{0.06} \\ \midrule
\multicolumn{1}{l|}{}                                                                                     & \begin{tabular}[c]{@{}c@{}}Intent \\ Uncertain\end{tabular} & \begin{tabular}[c]{@{}c@{}}Trajectory \\ Uncertain\end{tabular} & 1s   & 2s   & 3s   & \cellcolor[HTML]{C0C0C0}Avg. & 1s   & 2s   & 3s   & \cellcolor[HTML]{C0C0C0}Avg. \\
\multicolumn{1}{l|}{}                                                                                     & -                                                             & $\surd$                                                                 & 0.26 & 0.49 & 0.82 & \cellcolor[HTML]{C0C0C0}0.52 & 0.05 & 0.11 & 0.32 & \cellcolor[HTML]{C0C0C0}0.16 \\
\multicolumn{1}{l|}{}                                                                                     & $\surd$                                                             & -                                                                 & 0.22 & 0.41 & 0.69 & \cellcolor[HTML]{C0C0C0}\textbf{0.44} & 0.09 & 0.13 & 0.27 & \cellcolor[HTML]{C0C0C0}0.16 \\
\multicolumn{1}{c|}{\multirow{-4}{*}{\begin{tabular}[c]{@{}c@{}}Hierarchical \\ Planning\end{tabular}}}   & $\surd$                                                             & $\surd$                                                                 & 0.24 & 0.45 & 0.74 & \cellcolor[HTML]{C0C0C0}0.48 & 0.00 & 0.02 & 0.15 & \cellcolor[HTML]{C0C0C0}\textbf{0.06} \\ \midrule
\end{tabular}
\vspace{-3pt}
\end{table*}

\begin{table*}[!t]
\centering
\vspace{-5pt}
\caption{Ablation for the design of model details.}
\vspace{-5pt}
\label{tab:ablation4}
\begin{tabular}{cccllll|llll}
\toprule
                                                                        &                                                                          &                                                                              & \multicolumn{4}{c|}{L2 (m) $\downarrow$}                                 & \multicolumn{4}{c}{Collision (\%) $\downarrow$}                          \\
\multirow{-2}{*}{\begin{tabular}[c]{@{}c@{}}BEV\\ Adapter\end{tabular}} & \multirow{-2}{*}{\begin{tabular}[c]{@{}c@{}}Mode\\ Sharing\end{tabular}} & \multirow{-2}{*}{\begin{tabular}[c]{@{}c@{}}Ego-Motion\\ Interaction\end{tabular}} & 1s   & 2s   & 3s   & \cellcolor[HTML]{C0C0C0}Avg.          & 1s   & 2s   & 3s   & \cellcolor[HTML]{C0C0C0}Avg.          \\ \midrule
-                                                                       & $\surd$                                                                        & $\surd$                                                                            & 0.25 & 0.47 & 0.80 & \cellcolor[HTML]{C0C0C0}0.51          & 0.47 & 0.55 & 0.78 & \cellcolor[HTML]{C0C0C0}0.60          \\
$\surd$                                                                       & -                                                                        & $\surd$                                                                            & 0.22 & 0.41 & 0.68 & \cellcolor[HTML]{C0C0C0}\textbf{0.44} & 0.05 & 0.15 & 0.18 & \cellcolor[HTML]{C0C0C0}0.13          \\
$\surd$                                                                       & $\surd$                                                                        & -                                                                            & 0.25 & 0.46 & 0.77 & \cellcolor[HTML]{C0C0C0}0.49          & 0.05 & 0.10 & 0.25 & \cellcolor[HTML]{C0C0C0}0.13          \\
$\surd$                                                                       & $\surd$                                                                        & $\surd$                                                                            & 0.24 & 0.45 & 0.74 & \cellcolor[HTML]{C0C0C0}0.48          & 0.00 & 0.02 & 0.15 & \cellcolor[HTML]{C0C0C0}\textbf{0.06} \\ \midrule
\end{tabular}
\vspace*{-5mm}
\end{table*}

\paragraph{Open-loop Planning Results.} 
We compare CogAD against recent state-of-the-art (sota) end-to-end methods, including LLM-based planners. 
As shown in \cref{tab:performance}, CogAD demonstrates significant advantages in both performance and efficiency. 
Specifically, CogAD achieves a sota average collision rate of 0.06, substantially outperforming previous methods, including those enhanced with LLM. 
Meanwhile, CogAD maintains competitively low L2 error and high computational efficiency. 
To our knowledge, CogAD is the first to achieve 0\% collision rate at the 1s planning horizon without relying on ego-state or historical ego information. 
CogAD demonstrates superior overall performance in terms of L2 error, collision rate, and inference latency. 
Notably, this is achieved without employing resource-intensive techniques, such as the massive parameter counts or the post-selection refinement. 
These results highlight the effectiveness and efficiency of CogAD in real-world autonomous driving scenarios.

\paragraph{Long-tail Scenario Results.}
Following TOKEN~\cite{tian2024tokenize}, we construct a long-tail scenario validation set curated from nuScenes for comprehensive evaluation. 
As shown in \cref{tab:performance-long-tail}, CogAD achieves significant performance improvements over previous methods. 
Notably, CogAD establishes new sota results in L2 error for all three long-tail scenarios. 
In the challenging "3-point turn", a zero-shot case absent from training data, CogAD achieves a 10\% reduction in L2 error compared to previous sota result. 
For the "overtake" scenario, CogAD achieves a competitively low collision rate that nearly matches TOKEN~\cite{tian2024tokenize}'s performance, while simultaneously outperforming it in average L2 error metric by 0.18m.

\paragraph{Closed-loop Planning Results.} 
As shown in \cref{tab:b2d1}, the closed-loop evaluation of CogAD on the challenging Bench2Drive benchmark demonstrates its superior performance. 
Compared to other methods, CogAD achieves state-of-the-art performance in Driving Score and Success Rate, alongside competitive latency.

\subsection{Ablation Study}

\paragraph{Effect of Designs in Hierarchical Perception.} 
To validate the effectiveness of the hierarchical perception paradigm, we conduct two ablation studies: one retaining only the ego queries' interaction with BEV features, and the other preserving solely the ego-to-instance queries interaction. 
As shown in \cref{tab:ablation1}, interaction with BEV features benefits both the L2 error and collision rate, while instance-level interaction predominantly reduces the collision rate. 
These results collectively validate the effectiveness of our hierarchical perception paradigm.

\paragraph{Effect of Designs in Hierarchical Planning.}
\cref{tab:ablation1} demonstrates the effectiveness of our hierarchical planning design. 
The absence of intent uncertainty degrades CogAD's planning module to a conventional multi-mode ego-trajectory planning method, resulting in a significantly higher collision rate. 
Besides, preserving only intent uncertainty while omitting trajectory uncertainty also yields suboptimal results, as this configuration eliminates the model's capacity to generate refined planning trajectories. 
Our visualization analysis confirms that hierarchical planning delivers critical advantages in collision avoidance, enabling maneuvers like deceleration for yielding and lane-changing to bypass obstacles. This explains the results in \cref{tab:ablation1}: Collision Rate decreases by 62.5\% versus non-hierarchical planning counterparts, while L2 Error remains statistically comparable (showing minimal variation).

\begin{figure*}[!t]
  \centering
  \includegraphics[width=0.95\linewidth]{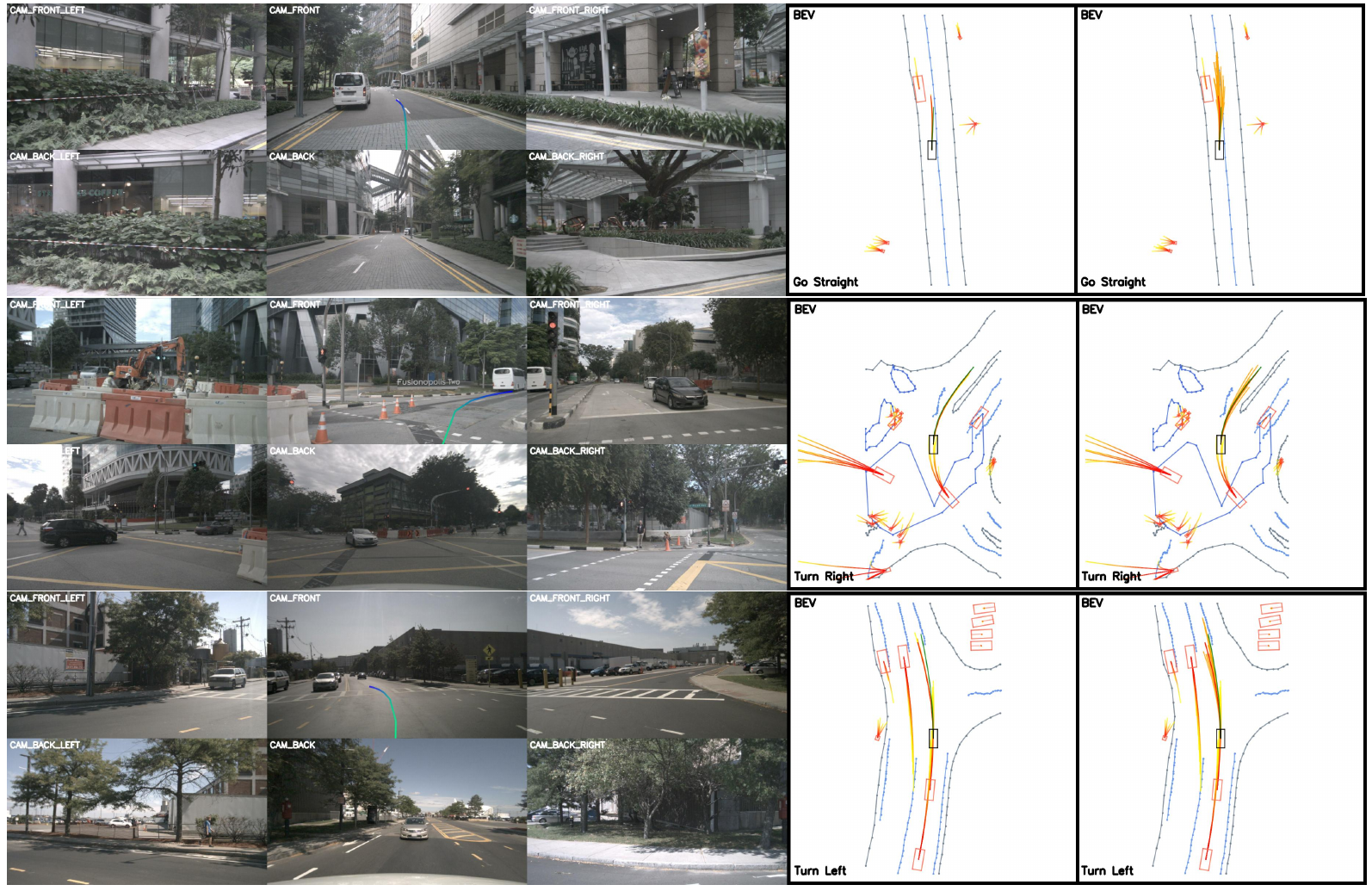}
  \caption{Qualitative results of CogAD on nuScenes. 
  The 2nd column shows the Top-3 multi-mode trajectories of the highest-probability intent, with colors of \textcolor{red}{red}, \textcolor{orange}{orange}, and \textcolor{yellow}{yellow}, respectively. 
  The 3rd column displays the Top-10 multi-intent trajectories with the highest-probability mode. 
  GT trajectory is drawn in \textcolor{green}{green}.}
  \label{fig:quality1}
  \vspace{-8pt}
\end{figure*}

\paragraph{Ablation for Model Design Details.} 
We validate the effectiveness of model design details, including the usage of BEV adapter, Mode Sharing, and Ego-Motion interaction. 
Mode Sharing indicates whether the ego query and motion query utilize distinct or shared trajectory mode embeddings. 
In the Ego-Motion interaction ablation, we remove the bidirectional interaction between ego query and motion query, retaining only the one-way interaction where ego query cross-attends motion query, which destroys the integrity of interactions between instances. 
As shown in \cref{tab:ablation4}, BEV adapter significantly improves the planning performance, which decreases the collision rate by nearly 90\%, demonstrating its clear advantage over using shared BEV features alone. 
The removal of either Mode Sharing or bidirectional Ego-Motion interaction maintains comparable L2 error in CogAD, but elevates the collision rate from 0.06\% to 0.13\% (116.6\% rise). 

\begin{figure}[!t]
  \centering
  \includegraphics[width=1.0\columnwidth]{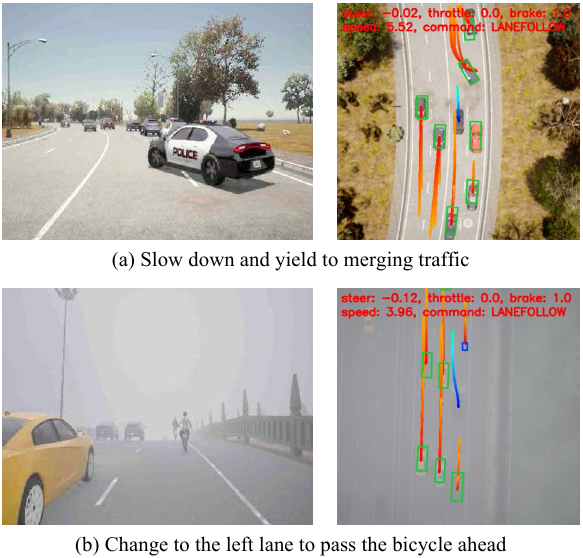}
  \caption{Qualitative results of CogAD on Bench2Drive.}
  \label{fig:b2d}
  \vspace{-10pt}
\end{figure}

\subsection{Qualitative Results} 
In \cref{fig:quality1}, we visualize the planning results of CogAD on nuScenes. 
Both the surrounding camera images and the prediction results in BEV space are provided accordingly. 
CogAD exhibits the capability to generate hierarchically dual-level multimodal trajectories. 
The rightmost column displays the highest-confidence trajectories corresponding to the top-10 intents, demonstrating that intent-level multimodal planning is diverse yet valid. 
The second column illustrates trajectory-level multi-mode planning conditioned on the dominant intent, with visualizations of the top-3 candidate trajectories. 
Multi-mode planning produces multiple refined trajectory options that maintain motion consistency. 
By combining multi-intent diversity and multi-mode refinement, CogAD generates multiple behaviorally diverse yet physically plausible planning candidates. 

Our qualitative analysis demonstrates that CogAD exhibits behaviors analogous to human cognition, as evidenced by the highest-probability planned trajectory visualized in \cref{fig:b2d}. 
These behaviors include strategic speed reduction for yielding, proactive lane changes for obstacle avoidance, and context-aware trajectory planning.

%% file: sec/5_conclusion.tex
\section{Conclusion}
We present CogAD, a hierarchical end-to-end planning method inspired by human driving processes through the lens of cognitive science.
For hierarchical perception, CogAD incorporates BEV-Instance interaction and cross-task instance interaction, which significantly enhance the ego vehicle's scene understanding capabilities. 
For hierarchical planning, CogAD utilizes intent anchors and trajectory modes to ensure behavioral diversity and geometric precision in generated trajectories. 
Extensive evaluations demonstrate that our approach achieves state-of-the-art performance across both open-loop and closed-loop benchmarks.

%% file: sec/X_suppl.tex
\begin{figure*}[t!]
  \centering
  \caption*{\LARGE \textbf{Supplementary Material}}
\end{figure*}


\setcounter{section}{0}
\renewcommand{\thesection}{\Alph{section}}

\section{Datasets}

\paragraph{Open-Loop.} 

nuScenes is a challenging public dataset for autonomous driving evaluation, which consists of 28k total samples in a 22k/6k training/validation split. The objects in each scene are annotated with 3D bounding box, orientation, and vehicle speed information. This dataset comprises 1000 complex driving scenarios, each spanning approximately 20 seconds, with annotations at a frame rate of 2Hz. For evaluation, we employ the L2 error and collision rate (CR) metrics following the evaluation protocol established in~\cite{jiang2023vad} to ensure fair comparison with recent sota works. 

\paragraph{Long-tail Scenario.}

We follow TOKEN~\cite{tian2024tokenize} and manually construct a long-tail scenario validation set curated from nuScenes for comprehensive evaluation, including three scenarios: 1) executing 3-point turns; 2) resuming
motion after a full stop; 3) overtaking parked cars through the oncoming lane. Details of each scenario are provided in \cref{tab:supply-long-tail}.

\paragraph{Closed-Loop.} 
Bench2Drive is a comprehensive evaluation protocol based on CARLA for evaluating abilities of end-to-end autonomous driving systems. 
We follow the standardized data partitioning and use 950 clips for training. 
For closed-loop evaluation, the benchmark provides 220 predefined short routes to assess dynamic planning capabilities in complex environments. 
Closed-loop evaluation employs five metrics: Driving Score (DS), Success Rate (SR), Efficiency, and Comfortness. Driving Score integrates route completion rate with multiplicative penalty factors for traffic violations. 
Success Rate quantifies the proportion of routes fully completed within predefined time constraints. 
Efficiency evaluates normalized speed performance relative to traffic compliance and route complexity. 
Comfortness measures ride smoothness through cumulative jerk and lateral acceleration integrals.

\begin{table}[t]
\caption{Long-tail.}
\label{tab:supply-long-tail}
\centering
\begin{tabular}{cllc}
\toprule
\multicolumn{1}{l}{Scenario}                  & Scene ID   & Frames Interval & \makecell{Frames \\ Number}        \\ \midrule
\multicolumn{1}{l}{3-point turn}              & scene-0778 & frame 6-30      & 25                   \\ \midrule
\multirow{8}{*}{\makecell{Resume \\from stop}}             & scene-0921 & frame 21-25     & \multirow{8}{*}{40}  \\
                                              & scene-0925 & frame 19-23     &                      \\
                                              & scene-0968 & frame 7-11      &                      \\
                                              & scene-0552 & frame 13-17     &                      \\
                                              & scene-0917 & frame 24-28     &                      \\
                                              & scene-0221 & frame 11-15     &                      \\
                                              & scene-1064 & frame 21-25     &                      \\
                                              & scene-0331 & frame 8-12      &                      \\ \midrule
\multicolumn{1}{l}{\multirow{5}{*}{Overtake}} & scene-0038 & frame 4-33      & \multirow{5}{*}{102} \\
\multicolumn{1}{l}{}                          & scene-0271 & frame 3-11      &                      \\
\multicolumn{1}{l}{}                          & scene-0969 & frame 14-33     &                      \\
\multicolumn{1}{l}{}                          & scene-0329 & frame 3-33      &                      \\
\multicolumn{1}{l}{}                          & scene-1065 & frame 24-35     &                      \\ \midrule
\end{tabular}
\end{table}
\vspace{5mm}

\begin{figure*}[!h]
  \centering
  \includegraphics[width=0.8\linewidth]{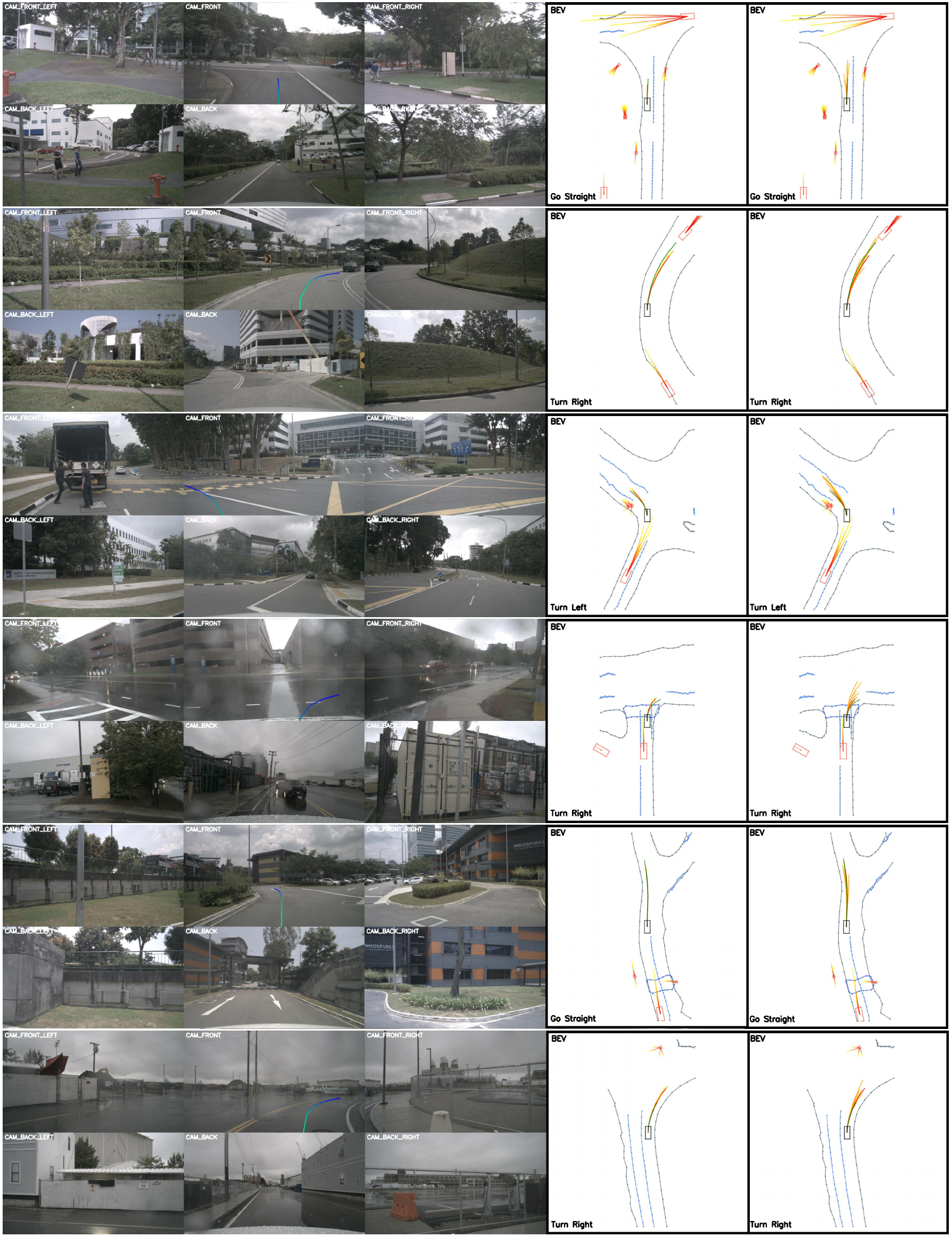}
    \caption{Qualitative results of CogAD on nuScenes. 
  The 2nd column shows the Top-3 multi-mode trajectories of the highest-probability intent, with colors of \textcolor{red}{red}, \textcolor{orange}{orange}, and \textcolor{yellow}{yellow}, respectively. 
  The 3rd column displays the Top-10 multi-intent trajectories with the highest-probability mode. 
  GT trajectory is drawn in \textcolor{green}{green}. 
  CogAD demonstrates robust performance across diverse commands, scenarios, and weather conditions.}
  \label{fig:supply1}
\end{figure*}

\begin{figure*}[!h]
  \centering
  \includegraphics[width=0.7\linewidth]{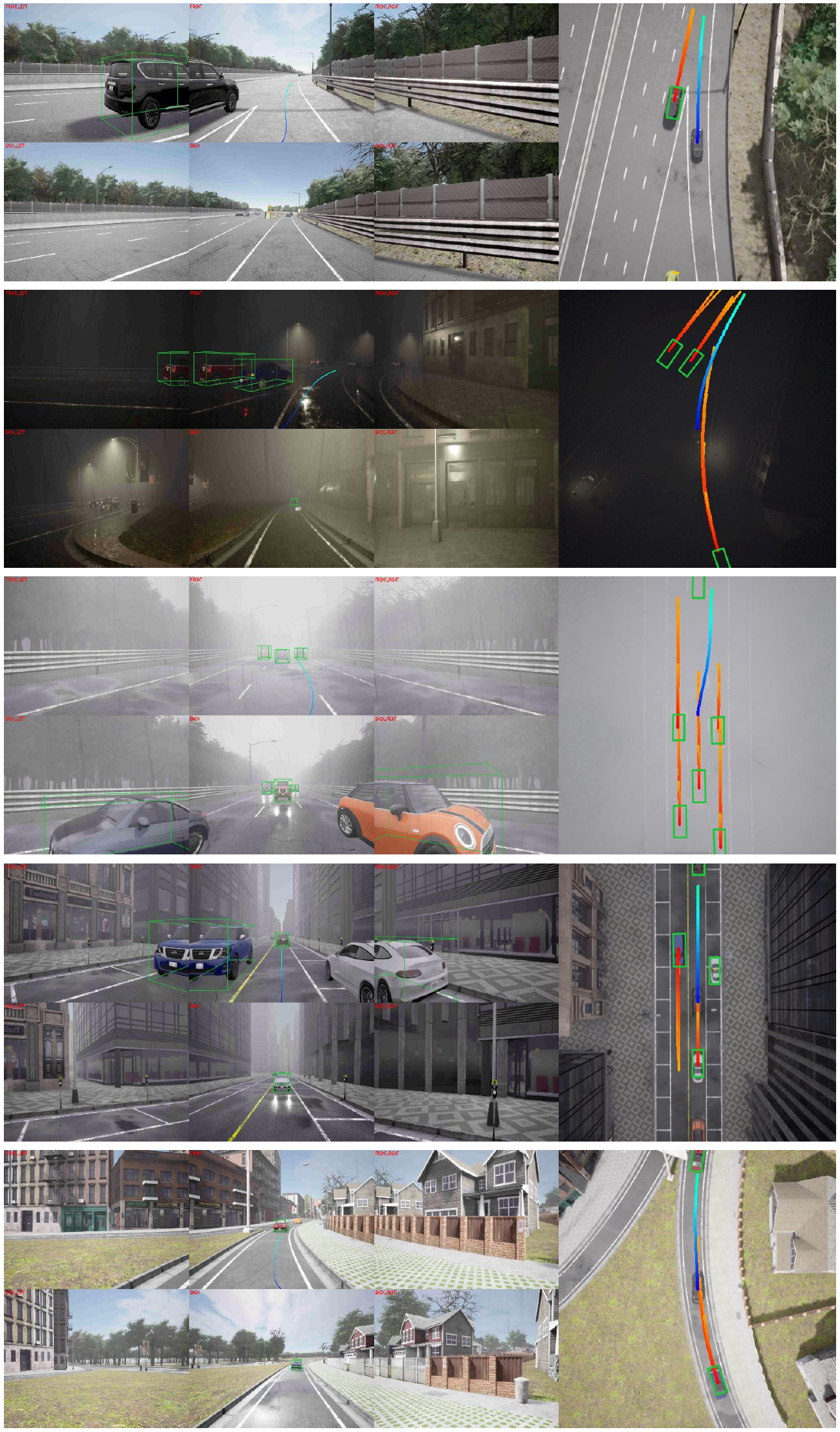}
  \caption{Qualitative results of CogAD on Bench2Drive closed-loop. 
  In the rightmost column, the centrally positioned black vehicle denotes the ego vehicle. 
  CogAD demonstrates robust performance across diverse commands, scenarios, and weather conditions.}
  \label{fig:supply2}
\end{figure*}

\begin{figure*}[!h]
  \centering
  \includegraphics[width=0.8\linewidth]{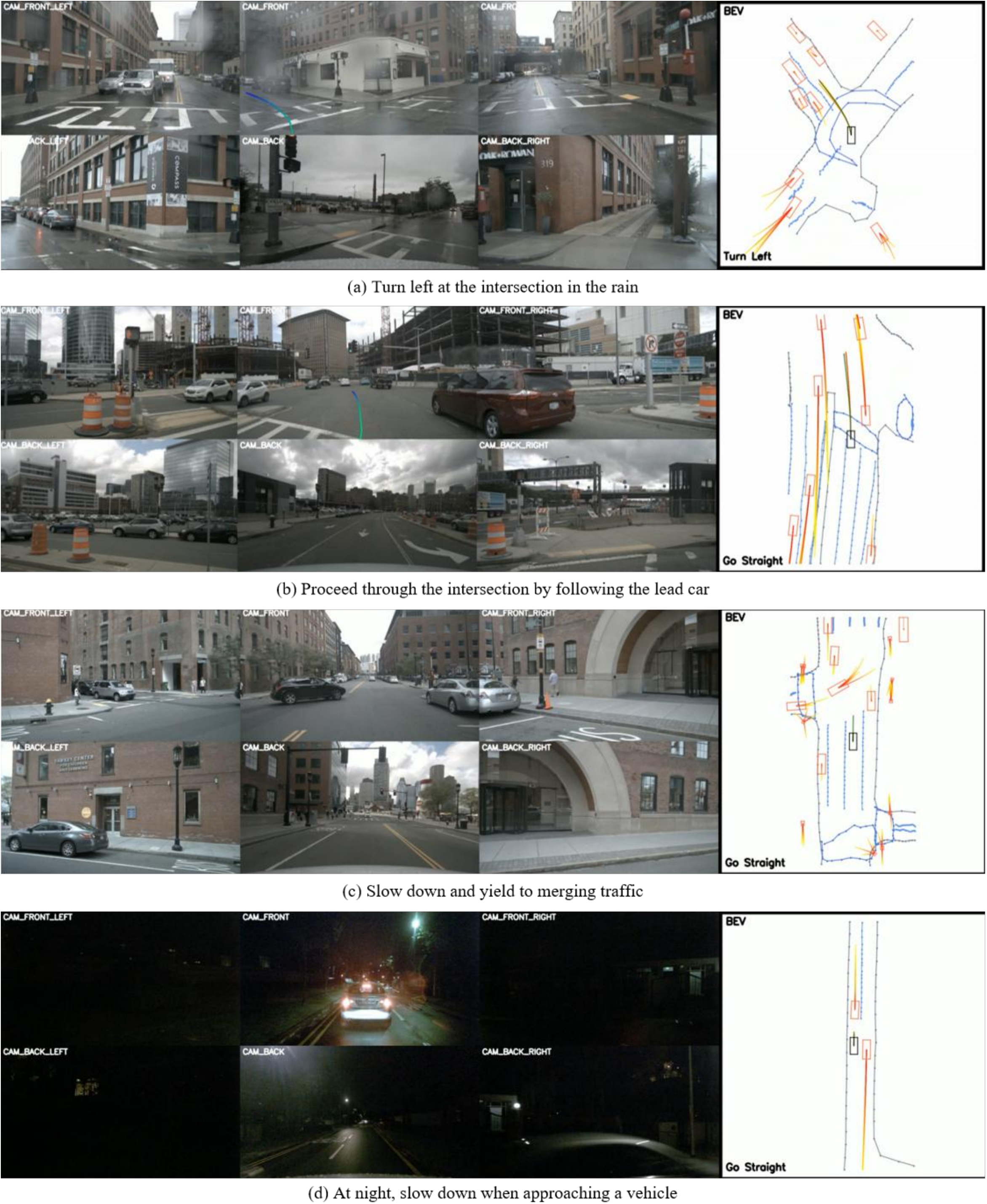}
    \caption{Qualitative results of CogAD on nuScenes. }
  \label{fig:hardcase_nus}
\end{figure*}

\begin{figure*}[!h]
  \centering
  \includegraphics[width=0.6\linewidth]{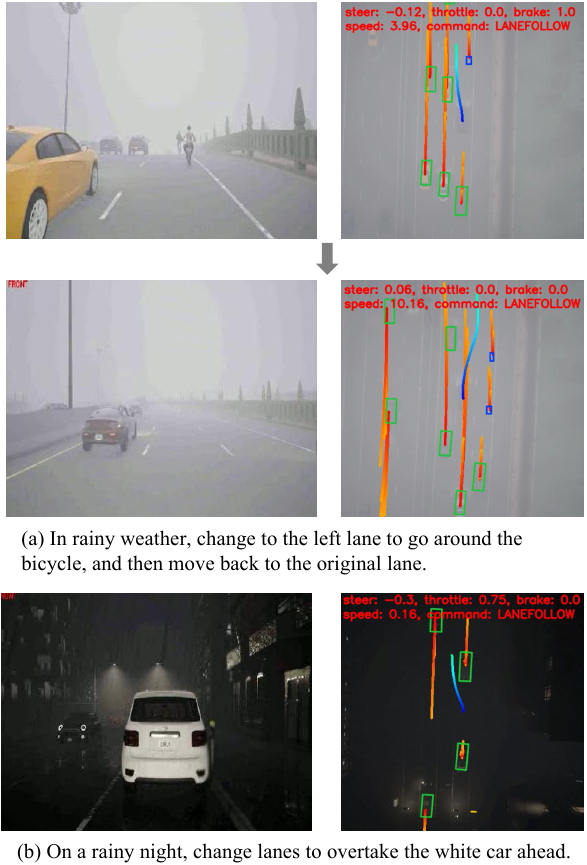}
    \caption{Qualitative results of CogAD on Bench2Drive. }
  \label{fig:hardcase_b2d}
\end{figure*}

\section{Implementation Details} 

CogAD plans a 3s future ego-trajectory without using any form of ego state or history information as input. 
The BEV perception range spans 60m longitudinally and 30m laterally, with input images resized to 640 × 360 pixels.  
We use a 100 × 100 BEV feature map, 100 × 20 map queries, and 300 agent queries. 
Moreover, we set the number of intent anchors to 30 and the trajectory modes to 6. 
The feature dimension size is set to 256. 
We train CogAD using AdamW optimizer~\cite{loshchilov2017decoupled} and Cosine Annealing scheduler~\cite{loshchilov2016sgdr} with initial learning rate $ 4 \times 10^{-4} $ and weight decay 0.01. 
CogAD is trained for 60 epochs on the nuScenes dataset and 6 epochs on the Bench2Drive dataset, utilizing 8 NVIDIA Tesla A100 GPUs with a total batch size of 32. 
\vspace{10mm}

\section{More Qualitative Results} 
 We provide more visualization results to illustrate the effectiveness of CogAD on various driving scenarios as shown in \cref{fig:supply1}, \cref{fig:supply2}, \cref{fig:hardcase_nus}, and \cref{fig:hardcase_b2d}.

